\documentclass{article}
\usepackage[utf8]{inputenc}
\usepackage{times}
\usepackage{bm}

\usepackage{natbib}
\usepackage{dblfloatfix}

\usepackage{xcolor}

\usepackage{graphicx}

\usepackage{hyperref}

\usepackage{flushend}

\usepackage[accepted]{icml2014_arxiv} 

\usepackage[fleqn]{amsmath} 
\usepackage{amsthm, amssymb}

\usepackage{mario-def}
\usepackage{parenthesis}

\newcommand{\cond}{\middle|}
\newcommand{\Acal}{\mathcal{A}}

\newcommand{\hStar}{h^{\star}}

\newcommand{\op}[1]{\operatorname{#1}}

\icmltitlerunning{Sequential Model-Based Ensemble Optimization}

\begin{document} 

\twocolumn[
\icmltitle{Sequential Model-Based Ensemble Optimization}

\icmlauthor{Alexandre Lacoste$^{*}$}{alexandre.lacoste.1@ulaval.ca}
\icmladdress{Département d’informatique et de génie logiciel,
            Université Laval, Québec, Canada, G1K-7P4}
\icmlauthor{Hugo Larochelle}{hugo.larochelle@usherbrooke.ca}
\icmladdress{Département d’informatique,
            Université de Sherbrooke, Québec, Canada, J1K-2R1}
\icmlauthor{Mario Marchand}{Mario.Marchand@ift.ulaval.ca}
\icmladdress{Département d’informatique et de génie logiciel,
            Université Laval, Québec, Canada, G1K-7P4}
\icmlauthor{François Laviolette}{Francois.Laviolette@ift.ulaval.ca}
\icmladdress{Département d’informatique et de génie logiciel,
            Université Laval, Québec, Canada, G1K-7P4}
\icmlkeywords{Machine Learning, Ensemble Learning, SMBO, Gaussian Process, Hyperparameter Optimization}

\vskip 0.3in
]


\begin{abstract} 
One of the most tedious tasks in the application
of machine learning is model selection, i.e.\
hyperparameter selection. Fortunately, recent
progress has been made in the automation of this
process, through the use of sequential model-based optimization (SMBO) methods. This can be used
to optimize a cross-validation performance of
a learning algorithm over the value of its hyperparameters. However, it is well known that
 ensembles of learned models
almost consistently outperform a single model, even
if properly selected. In this paper, we thus
propose an extension of SMBO methods that automatically constructs such ensembles. This method builds on a recently proposed ensemble construction paradigm known as agnostic Bayesian learning. In experiments
on 22 regression and 39 classification data sets, we confirm the success of this proposed approach, which is able to outperform model selection with SMBO. 
\end{abstract}



\section{Introduction}

The automation of hyperparameter selection is an important step towards making the practice
of machine learning more approachable to the non-expert and increases its impact on data reliant sciences. Significant progress has been made recently, with many methods reporting success in tuning a large variety of algorithms~\cite{bergstra2011,hutter2011sequential,snoek2012practical}. One successful general paradigm
is known as Sequential Model-Based Optimization (SMBO). It is based on a process that alternates
between the proposal of a new hyperparameter configuration to test and the update of an adaptive model
of the relationship between hyperparameter configurations and their holdout set performances. Thus, as the model learns about this relationship, it increases its ability to suggest
improved hyperparameter configurations and gradually converges to the best solution.

While finding the single best model configuration is useful, better performance
is often obtained by, instead, combining several (good) models into an ensemble. This was best illustrated by the winning entry of the Netflix competition, which combined a variety
of models~\citep{bell2007bellkor}. Even if one concentrates on a single learning algorithm, combining models
produced by using different hyperparameters is also helpful.
Intuitively, models with comparable performances are still likely to generalize differently across the input space and produce different patterns of errors. By averaging their predictions, we can hope that the majority of models actually perform well on any given input and will move the ensemble towards better predictions globally, by dominating the average. In other words, the averaging of several comparable models reduces the variance of our predictor compared to each individual in the ensemble, while not sacrificing too much in terms of bias.

However, constructing such ensembles is just as tedious as performing model selection and at least as important in the successful deployment of machine-learning-based systems. Moreover, unlike the model selection case for which SMBO can be used, no comparable automatic ensemble construction methods have been developed thus far. The current methods of choice remain trial and error or exhaustive grid search for exploring the space of models to combine, followed by a selection or weighting strategy which is often an heuristic.

In this paper, we propose a method for leveraging the recent research on SMBO in order to generate an ensemble of models, as opposed to the single best model. The proposed approach builds on the agnostic Bayes framework~\citep{agnostic_bayes}, which provides a successful strategy for weighting a predetermined and finite set of models (already trained) into an ensemble. Using a successful SMBO method, we show how we can effectively generalize this framework to the case of an infinite space of models (indexed by its hyperparameter space). The resulting method is simple and highly efficient. Our experiments on 22 regression and 39 classification data sets confirm that it outperforms the regular SMBO model selection method.

The paper develops as follows. First, we describe SMBO and its use for hyperparameter selection (Section~\ref{sec:smbo}). We follow with a description of the agnostic Bayes framework and present a bootstrap-based implementation of it (Section~\ref{sec:ab}). Then, we describe the proposed algorithm for automatically constructing an ensemble using SMBO (Section~\ref{sec:our_method}). Finally, related work is discussed (Section~\ref{sec:related}) and the experimental comparisons
are presented (Section~\ref{sec:experiments}).

\section{Hyperparameter selection with SMBO}
\label{sec:smbo}

Let us first lay down the notation we will be using to describe
the task of model selection for a machine learning algorithm. In this setup, a task $D$
corresponds to a probability distribution over the input-output space
$\Xcal \times \Ycal$. Given a set of examples $S\sim D^m$ (which will be our holdout validation set), the objective
is to find, among a set $\Hcal$, the \emph{best} function $\hStar:
\Xcal \to \Ycal$. In general, $\Hcal$ can be any set and we refer to a member as a predictor. In the context of hyperparameter selection, $\Hcal$ corresponds to the set
of models trained on a training set $T\sim D^n$ (disjoint from $S$), for different configurations of the learning algorithm's hyperparameters $\gamma$. Namely, let $\Acal_{\gamma}$ be the learning algorithm with a hyperparameter configuration $\gamma \in \Gamma$,
we will note $h_{\gamma} = \Acal_{\gamma}(T)$ the predictor obtained after training on $T$ . The set $\Hcal$ contains all
predictors obtained from each $\gamma  \in \Gamma$ when $\Acal_{\gamma}$ is trained on $T$, i.e. $\Hcal \eqdef \l\{h_{\gamma} \cond \gamma \in \Gamma \r\}$. 

To assess the quality of a predictor, we use a loss function
$\Lcal:\Ycal \times \Ycal \to \Reals$ \,that quantifies the penalty
incurred when $h_{\gamma}$ predicts $h_\gamma(x)$ while the true target is $y$. Then,
we can define the risk $R_D(h_{\gamma})$ as being the expected loss of $h_{\gamma}$ on
task $D$, i.e.\ $R_D(h_{\gamma}) \eqdef \esp{x,y \sim D} \left[\Lcal\l( h_{\gamma}(x),y \r)\right]
$. Finally, the \emph{best}\footnote{The best solution may not be 
  unique but any of them are equally good.} function is simply the one minimizing the risk, i.e.\
$\hStar \eqdef \argmin{h_{\gamma} \in \Hcal} R_D(h_{\gamma}) $. Here, estimating $\hStar$ thus corresponds to hyperparameter selection.

For most of machine learning history, the state of the art in hyperparameter selection has been testing a list of predefined configurations and selecting the best according to the loss function $\Lcal$ on some holdout set of examples $S$. When a learning algorithm has more than one hyperparameter, a grid search is required, forcing $|\Gamma|$ to grow exponentially with the number of hyperparameters. In addition, the search may yield a suboptimal result when the minimum lies outside of the grid or when there is not enough computational power for an appropriate grid resolution. Recently,
randomized search has been advocated as a better replacement to grid search~\citep{bergstra2012randomhp}. While it tends to be superior to grid search, it remains inefficient since its search is not informed by results of the sequence of hyperparameters that are tested.

To address these limitations, there has been an increasing amount of work on automatic hyperparameter optimization~\citep{bergstra2011,hutter2011sequential,snoek2012practical}. Most rely on an approach called sequential  based  optimization (SMBO). The idea consists in treating $R_S(h_\gamma)\eqdef f(\gamma)$ as a learnable function of $\gamma$, which we can learn from the observations $\{(\gamma_i,R_S(h_{\gamma_i}))\}$ collected during the hyperparameter selection process. 

We must thus choose a model family for $f$. A common choice is a Gaussian process (GP) representation, which allows us to represent our uncertainty about $f$, i.e.\ our uncertainty about the value of $f(\gamma^*)$ at any unobserved hyperparameter configuration $\gamma^*$. This uncertainty can then be leveraged to determine an \emph{acquisition function} that suggests the
most promising hyperparameter configuration to test next. 

Namely, let functions $\mu : \Gamma \rightarrow \mathbb{R}$ and $K : \Gamma \times \Gamma \rightarrow \mathbb{R}$ be the mean and covariance kernel functions of our GP over $f$. Let us also denote the set of the $M$ previous evaluations as
\begin{equation}
\Rcal \eqdef \l\{ \l( \gamma_i, R_S\l(h_{\gamma_i}\r) \r) \r\}_{i=1}^M
\end{equation}
where $R_S\l(h_{\gamma_i}\r)$ is the empirical risk of $h_{\gamma_i}$ on set $S$, i.e.\ the holdout set error for hyperparameter $\gamma$.

The GP assumption on $f$ implies that the conditional distribution $p(f(\gamma^*)|\Rcal)$ is Gaussian, that is
\begin{eqnarray*}
&&p(f(\gamma^*)|\Rcal) = {\cal N}(f(\gamma^*);\mu(\gamma^*;\Rcal),\sigma^2(\gamma^*;\Rcal),\\
&&\mu(\gamma^*;\Rcal) \eqdef \mu(\gamma^*) + {\bf k}^{\top}{\bf K}^{-1}({\bf r}-{\boldsymbol \mu}),\\
&&\sigma^2(\gamma^*;\Rcal) \eqdef K(\gamma^*,\gamma^*) - {\bf k}^{\top}{\bf K}^{-1}{\bf k}
\end{eqnarray*}
where ${\cal N}(f(\gamma^*);\mu(\gamma^*;\Rcal),\sigma^2(\gamma^*;\Rcal)$ is the Gaussian density function with mean $\mu(\gamma^*;\Rcal)$ and variance $\sigma^2(\gamma^*;\Rcal)$. We also have vectors ${\boldsymbol \mu} \eqdef [\mu(\gamma_1),\dots,\mu(\gamma_M)]^\top$, ${\bf k} \eqdef [K(\gamma^*,\gamma_1),\dots,K(\gamma^*,\gamma_M)]^\top$, ${\bf r} \eqdef [R_S\l(h_{\gamma_1}\r) , \dots,R_S\l(h_{\gamma_M}\r) ]^\top$, and matrix ${\bf K}$ is such that ${\bf K}_{ij} = K(\gamma_i,\gamma_j)$. 

There are several choices for the acquisition function. One that has been used with success is the one maximizing the \emph{expected improvement}:
\begin{equation}
{\rm EI}(\gamma^*;\Rcal) \eqdef{\rm E}\left[\max\{r_{\rm best}-f(\gamma^*),0\}|\Rcal\right]
\end{equation}
which can be shown to be equal to 
\begin{equation}
\sigma^2(\gamma^*;\Rcal) \left(d(\gamma^*;\Rcal)\Phi(d(\gamma^*;\Rcal))+ {\cal N}(d(\gamma^*;\Rcal),0,1)\right)\label{eq:EI}
\end{equation}
where $\Phi$ is the cumulative distribution function of the standard normal and
\begin{eqnarray*}
r_{\rm best} &\eqdef& \min_i R_S\l(h_{\gamma_i}\r),\\
d(\gamma^*;\Rcal) &\eqdef& \frac{r_{\rm best}-\mu(\gamma^*;\Rcal)}{\sigma(\gamma^*;\Rcal)}~.
\end{eqnarray*}
The acquisition function thus maximizes Equation~\ref{eq:EI} and returns its solution. This optimization can be performed by gradient ascent initialized at points distributed across the hyperparameter space according to a Sobol sequence, in order to maximize the chance of finding a global optima. One advantage of expected improvement is that it directly offers a solution to the exploration-exploitation trade-off that hyperparameter selection faces.

An iteration of SMBO requires fitting the GP to the current set of tested hyperparameters $\Rcal$ (initially empty), invoking the acquisition function, running the learning algorithm with the suggested hyperparameters and adding the result to $\Rcal$. This procedure is expressed in Algorithm~\ref{alg:fast-hp}. Fitting the GP corresponds to learning the mean and covariance functions hyperparameters to the collected data. This can be performed either by maximizing the data's marginal likelihood or defining priors over the hyperparameters and sampling from the posterior using  sampling (see \citet{snoek2012practical} for more details).

\begin{algorithm}
\caption{SMBO hyperparameter optimization with GPs}
\label{alg:fast-hp}
\begin{algorithmic}
\STATE $\Rcal \leftarrow \{ \}$  
\FOR{$k \in \{1,2,\dots,M\}$}
  \STATE $\gamma \leftarrow \op{SMBO}( \Rcal )$ \COMMENT{Fit GP and maximize EI}
  \STATE $h_{\gamma} \leftarrow \Acal_{\gamma}(T)$ \COMMENT{Train with suggested $\gamma$}
  \STATE $\Rcal \leftarrow \Rcal \cup \l\{ \l( \gamma, R_{S}(h_{\gamma}) \r) \r\}$ \COMMENT{Add to collected data}
\ENDFOR
\STATE $\gamma^* \leftarrow \argmin{ (\gamma, R_{S}(h_{\gamma})) \in \Rcal} R_{S}(h_{\gamma})$
\RETURN $h_{\gamma^*}$ 
\end{algorithmic}
\end{algorithm}

While SMBO hyperparameter optimization can produce very good predictors, it can also suffer from overfitting on the validation set, especially for high-dimensional hyperparameter spaces. This is in part why an ensemble of predictors are often preferable in practice. Properly extending SMBO to the construction of ensembles is, however, not obvious. Here, we propose one such successful extension, building on the framework of Agnostic Bayes learning, described in the next section.

\section{Agnostic Bayes} 
\label{sec:ab}

In this section, we offer a brief overview of the Agnostic Bayes learning paradigm presented in \citet{agnostic_bayes} and serving as a basis for the algorithm we present in this paper. Agnostic Bayes learning was used in~\citet{agnostic_bayes}
as a framework for successfully constructing ensembles when the number of predictors in $\Hcal$ (i.e.\ the potential hyperparameter configurations $\Gamma$) was constrained to be finite (e.g.\ by restricting the space to a grid). In our context, we can thus enumerate the possible hyperparameter configurations from $\gamma_1$ to $\gamma_{|\Gamma|}$. This paper will generalize this approach to the infinite case later.

Agnostic Bayes learning attempts to directly address the problem of inferring what is the  \emph{best} function $\hStar$ in $\Hcal$, according to the loss function $\Lcal$. It infers a posterior $p_{\hStar}(h_{\gamma} |S)$, i.e.\ a distribution over how likely each member of $\Hcal$ is the best predictor. This is in contrast with standard Bayesian learning, which implicitly assumes that $\Hcal$ contains the true data-generating model and infers a distribution for how likely each member of $\Hcal$ has generated the data (irrespective of what the loss $\Lcal$ is). 
From $p_{\hStar}(h_{\gamma} |S)$, by marginalizing $\hStar$ and selecting the most probable prediction, we obtain the following ensemble decision rule:
\begin{equation}\label{eqn:ensemble_decision}
E^\star(x) \eqdef \argmax{y \in \Ycal} \sum_{\gamma\in \Gamma} p_{\hStar}(h_{\gamma}|S) I[ h_\gamma(x) =y ].
\end{equation}

To estimate $p_{\hStar}(h_{\gamma} |S)$, Agnostic Bayes learning uses the set of losses $l_{\gamma,i} \eqdef \Lcal(h_\gamma(x_i), y_i)$ of each example $(x_i,y_i)\in S$ as evidence for inference. In \citet{agnostic_bayes}, a few different approaches are proposed and analyzed. A general strategy is to assume a joint prior $p({\bf r})$ over the risks $r_{\gamma} \eqdef R_D(h_{\gamma})$ of all possible hyperparameter configurations and choose a joint observation $p(l_{\gamma,i}~\forall\gamma \in \Gamma|{\bf r})$ for the losses.
From Bayes rule, we obtain the posterior $p({\bf r}|S)$ from which we can compute
\begin{eqnarray}
p_{\hStar}(h_{\gamma} |S) & = & {\rm E}_{{\bf r}}\left[I[r_{\gamma} < r_{\gamma'}, \forall \gamma' \neq \gamma]|S\right]\label{eq:posterior}
\end{eqnarray}
with a Monte Carlo estimate. This would result in repeatedly sampling from $p({\bf r}|S)$ and counting the number of times each $\gamma$ has the smallest sampled risk $r_\gamma$ to estimate $p_{\hStar}(h_{\gamma} |S)$. Similarly, samples from $p_{\hStar}(h_{\gamma} |S)$ could be obtained by sampling
a risk vector ${\bf r}$ from $p({\bf r}|S)$ and returning the predictor $h_\gamma$
with the lowest sampled risk. The ensemble decision rule of Equation~\ref{eqn:ensemble_decision} could then be implemented by repeatedly sampling from $p_{\hStar}( h_{\gamma} |S)$ to construct the ensemble of predictors and using their average as the ensemble's prediction.

Among the methods explored in \citet{agnostic_bayes} to obtain samples from $p({\bf r}|S)$, the bootstrap approach stands out for its efficiency and simplicity. Namely, to obtain a sample from $p({\bf r}|S)$, we sample with replacement from $S$ to obtain $S'$ and return the vector of empirical risks $[R_{S'}(h_{\gamma_1}),\dots,R_{S'}(h_{\gamma_{|\Gamma|}})]^\top$ as a sample. While bootstrap only serves as a "poor man's" posterior, it can be shown to be statistically related to a proper model with Dirichlet priors and its empirical performance was shown to be equivalent \citep{agnostic_bayes}. 

When the bootstrap method is used to obtain samples from $p_{\hStar}(h_{\gamma} |S)$, the complete procedure for generating each ensemble member can be summarized by
\begin{equation}
 \widetilde{\hStar} = \argmin{\gamma \in \Gamma } R_{S'}(h_\gamma) \label{eq:minboostrap}
\end{equation}
where $\widetilde{\hStar}$ is a returned sample.


\section{Agnostic Bayes ensemble with SMBO}
\label{sec:our_method}

We now present our proposed method for automatically constructing an ensemble,
without having to restrict $\Gamma$ (or, equivalently $\Hcal$) to a finite subset of hyperparameters. 

As described in Section~\ref{sec:ab}, to sample a predictor from the Agnostic Bayes bootstrap method, it suffices to obtain a bootstrap $S'$ from $S$ and solve the optimization problem of Equation~\ref{eq:minboostrap}. In our context where $\Hcal$ is possibly an infinite set
of models trained on the training set $T$ for any hyperparameter configuration $\gamma$, Equation~\ref{eq:minboostrap} corresponds in fact to hyperparameter optimization where the holdout set is $S'$ instead of $S$.

This suggests a simple procedure for building an ensemble of $N$ predictors according to agnostic Bayes i.e.,\ that reflects our uncertainty about the true best model $\hStar$. We could repeat the full SMBO hyperparameter optimization process $N$ times, with different bootstrap $S'_j$, for $j \in \{ 1,2,\dots, N\}$. However, for large ensembles, performing $N$ runs of SMBO can be computationally expensive, since each run would need to train its own sequence of models. 


We can notice however that predictors are always trained on the same training set $T$, no matter in which run of SMBO they were trained on. We propose a handy trick that exploits this observation to greatly accelerate the construction of the ensemble by almost a factor of $N$. Specifically, we propose to simultaneously optimize all $N$ problems in a round-robin fashion. Thus, we maintain $N$ different histories of evaluation $\Rcal_j$, for $j \in \{1,2,\dots,N\}$ and when a new predictor $h_\gamma = \Acal_{\gamma}(T)$ is obtained, we update all $\Rcal_j$ with $(\gamma, R_{S'_j}(h_{\gamma}))$. Notice that the different histories $\Rcal_j$ contain the empirical risks on different bootstrap holdout sets, but they are all updated at the cost of training only a single predictor. Also, to avoid recalculating multiple times $\Lcal( h_\gamma(x_i), y_i )$, these values can be cached and shared in the computation of each $\Rcal_j$. This leaves the task of updating all $\Rcal_j$ insignificant compared to the computational time usually required for training a predictor. This procedure is detailed in Algorithm~\ref{alg:fast-hp+ab}.
\begin{algorithm}
\caption{Agnostic Bayes Ensemble With SMBO}
\label{alg:fast-hp+ab}
\begin{algorithmic}
\STATE ${\cal E} \leftarrow \{\}$
\FOR{ $j \in \{1,2,\dots,N\}$}
  \STATE $\Rcal_j \leftarrow \{ \}$  
  \STATE $S'_j \leftarrow  \op{bootstrap}(S)$ 
\ENDFOR
\FOR{$k \in \{1,2,\dots,M\}$}
  \STATE $v \leftarrow k \op{modulo} N$ 
  \STATE $\gamma \leftarrow \op{SMBO}( \Rcal_v )$ 
  \STATE $h_\gamma \leftarrow \Acal_{\gamma}(T)$ 
  \STATE ${\cal E} \leftarrow {\cal E} \cup \{h_\gamma\}$
  \FOR{$j \in \{1,2,\dots,N\}$} 
    \STATE $\Rcal_j \leftarrow \Rcal_j \cup \l\{ \l( \gamma, R_{S'_j}(h_\gamma) \r) \r\}$  
  \ENDFOR
\ENDFOR
\RETURN ${\cal E}$ 
\end{algorithmic}
\end{algorithm}

By updating all $\Rcal_j$ at the same time, we \emph{trick} each SMBO run by updating its history with points it did not suggest. This implies that the GP model behind
each SMBO run will be able to condition on more observations then it would if the runs had been performed in isolation. This can only benefit the GPs and improve the quality of their suggestions. 

\section{Related Work}
\label{sec:related}

In the Bayesian learning literature, a common way of
dealing with hyperparameters in probabilistic predictors is to define
hyperpriors and perform posterior inference to integrate them out.
This process often results in also constructing an ensemble
of predictors with different hyperparameters, sampled from the posterior. 
Powerful MCMC methods have been developed in order to accommodate
for different types of hyperparameter spaces, including infinite spaces.

However, this approach requires that the family predictors in question be
probabilistic in order to apply Bayes rule. Moreover, even if the
predictor family is probabilistic, the construction of the ensemble will
entirely ignore the nature of the loss function that determines the
measure of performance. The comparative advantage of the proposed Agnostic Bayes SMBO
approach is thus that it can be used for any predictor family (probabilistic
or not) and is loss-sensitive.

On the other hand, traditional ensemble methods such as \citet{laviolette2011pac}, \citet{KimG12}, and \cite{zhang2006ensemble} require a predefined set of models and are not straightforward to adapt to an infinite set of models. 

\section{Experiments}
\label{sec:experiments}

We now compare the SMBO ensemble approach (ESMBO) to three alternative methods for building a predictor from a machine learning algorithm with hyperparameters:
\begin{itemize}
\item A single model, whose hyperparameters were selected by hyperparameter optimization with SMBO (SMBO). 
\item A single model, whose hyperparameters were selected by a randomized search (RS), which in practice is often superior to grid search~\citep{bergstra2012randomhp}.
\item An Agnostic Bayes ensemble constructed from a randomly selected set of hyperparameters (ERS). 
\end{itemize}
Both ESMBO and SMBO used GP models of the holdout risk, with hyperparameters
trained to maximize the marginal likelihood. A constant was used for the mean function, while the Mat\'ern~5/2 kernel was used for the covariance function, with length scale parameters. The GP's parameters were obtained by maximizing the marginal likelihood and a different length scale was used for each dimension\footnote{We used the implementation provided by spearmint: \url{https://github.com/JasperSnoek/spearmint}}.

Each method is allowed to evaluate 150 hyperparameter configurations. To compare their performances, we perform statistical tests on several different hyperparameter spaces over two different collections of data sets. 


\subsection{Hyperparameter Spaces}
Here, we describe the hyperparameter spaces of all learning algorithms we employ in our experiments.
Except for a custom implementation of the multilayer perceptron, we used scikit-learn\footnote{\url{http://scikit-learn.org/}} for the implementation of all other learning algorithms.

\paragraph{Support Vector Machine} We explore the soft margin parameter $C$ for values ranging from $10^{-2}$ to $10^3$ on a logarithmic scale. We use the RBF kernel $K(x,x') = e^{\gamma ||x-x'||_2^2}$ and explore values of $\gamma$ ranging from $10^{-5}$ to $10^3$ on a logarithmic scale. 

\paragraph{Support Vector Regressor} We also use the RBF kernel and we explore the same values as for the Support Vector Machine. In addition, we explore the $\epsilon$-tube parameter \citep{drucker1997support} for values ranging between $10^{-2}$ and $1$ on a logarithmic scale. 

\paragraph{Random Forest} We fix the number of trees to 100 and we explore two different ways of producing them: either the original \citet{breiman2001random} method or the extremely randomized trees method of \citet{geurts2006extremely}. We also explore the choice of bootstrapping or not the training set before generating a tree. Finally, the ratio of randomly considered features at each split for the construction of the trees is varied between $10^{-4}$ and $1$ on a linear scale. 

\paragraph{Gradient Boosted Classifier} This is a tree-based algorithm using boosting \citep{friedman2001greedy}. We fix the set of weak learners to 100 trees and take the maximum depth of each tree to be in $\{1,2,\dots,15\}$. The learning rate ranges between $10^{-2}$ and $1$ on a logarithmic scale.  Finally, the ratio of randomly considered features at each split for the construction of the trees varies between $10^{-3}$ and $1$ on a linear scale. 

\paragraph{Gradient Boosted Regressor} We use the same parameters as for Gradient Boosted Classifier except that we explore a convex combination of the least square loss function and the least absolute deviation loss function. We also fix the ratio of considered features at each split to 1.

\paragraph{Multilayer Perceptron} We use a 2 hidden layers perceptron with tanh activation function and a softmax function on the last layer. We minimize the negative log likelihood using the L-BFGS algorithm. Thus there is no learning rate parameter. However, we used a different L2 regularizer weight for each of the 3 layers with values ranging from $10^{-5}$ to $100$ on a logarithmic scale. Also, the number of neurons on each layer can take values in $\{1,2,\dots,100\}$. In total, this yields a 5 dimensional hyperparameter space.

\newcommand{\rank}{\operatorname{Rank}}

\subsection{Comparing Methods On Multiple Data Sets}\label{sec:context}

\newcommand{\better}[1]{\overset{{}_{#1}}{\succ}}

The different methods presented in this paper are generic and are
meant to work across different tasks. It is thus crucial that we evaluate them on several data sets using metrics that do not assume commensurability across tasks \citep{demvsar2006statistical}. The metrics of choice are thus the expected rank and the pairwise winning frequency. 
Let $\Acal_i(T_j,S_j)$ be either one of our $K=4$ model selection/ensemble construction algorithms run on the $j^{\rm th}$ data set, with training set $T_j$ and validation set $S_j$.
When comparing $K$ algorithms, the rank of 
(best or ensemble) predictor $h_i = \Acal_i(T_j,S_j)$ on test set $S^{\rm test}_j$ is defined as
\[
\rank_{h_i,S^{\rm test}} \eqdef \sum_{l=1}^{K} I\l[ R_{S^{\rm test}_j}(h_l) \leq R_{S^{\rm test}_j}(h_i) \r].
\]
Then, the expected rank of the $i^{th}$ method 
is obtained from the empirical average over the $L$ data sets i.e., $ \esp{}[\rank]_{i} \eqdef \frac{1}{L}  \sum_{j=1}^L \rank_{h_i,S^{\rm test}_j}$.
When comparing algorithm $\Acal_i$ against algorithm $\Acal_l$, the winning frequency\footnote{We deal with ties by attributing 0.5 to each method except for the Sign test where the sample is simply discarded.} of $\Acal_i$ is  
\[
\rho_{i,l} \eqdef \frac{1}{L} \sum_{i=1}^L{I[ R_{S^{\rm test}_j}(h_i) < R_{S^{\rm test}_j}(h_l) ]}
\]
In the case of the expected rank, lower is better and for the winning frequency, it is the converse. Also, when $K=2$, $\esp{}[\rank]_{i} = 1 + (1-\rho_{i,l})$.

When the winning frequency $\rho_{i,l} > 0.5$, we say that method ${\cal A}_i$ is better than method ${\cal A}_l$. However, to make sure that this is not the outcome of chance, we use statistical tests such as the sign test and the Poisson Binomial test (PB test) \citep{lacostebayesian}. The PB test derives a posterior distribution over $\rho_{i,l}$ and integrates the probability mass above $0.5$, denoted as $\Pr( A \better{} B )$. When $\Pr( A \better{} B ) > 0.8$, we say that the result is significant and when $\Pr( A \better{} B ) > 0.9$, we say that it is highly significant. Similarly for the sign test, when the $p$-value is lower than 0.1, it is significant and when lower than $0.05$, it is highly significant.

To build a substantial collection of data sets, we used the
AYSU collection \citep{ulas2009} coming from the UCI and the Delve
repositories and we added the MNIST data set.
We also converted the multiclass data sets to binary classification by either merging
classes or selecting pairs of classes. The resulting benchmark contains 39
data sets. We have also collected 22 regression data sets from the Louis Torgo collection\footnote{These data sets were obtained from the following source : \url{http://www.dcc.fc.up.pt/~ltorgo/Regression/DataSets.html}}.

\begin{figure*}[t]
\begin{center}
\includegraphics[width=0.9\textwidth]{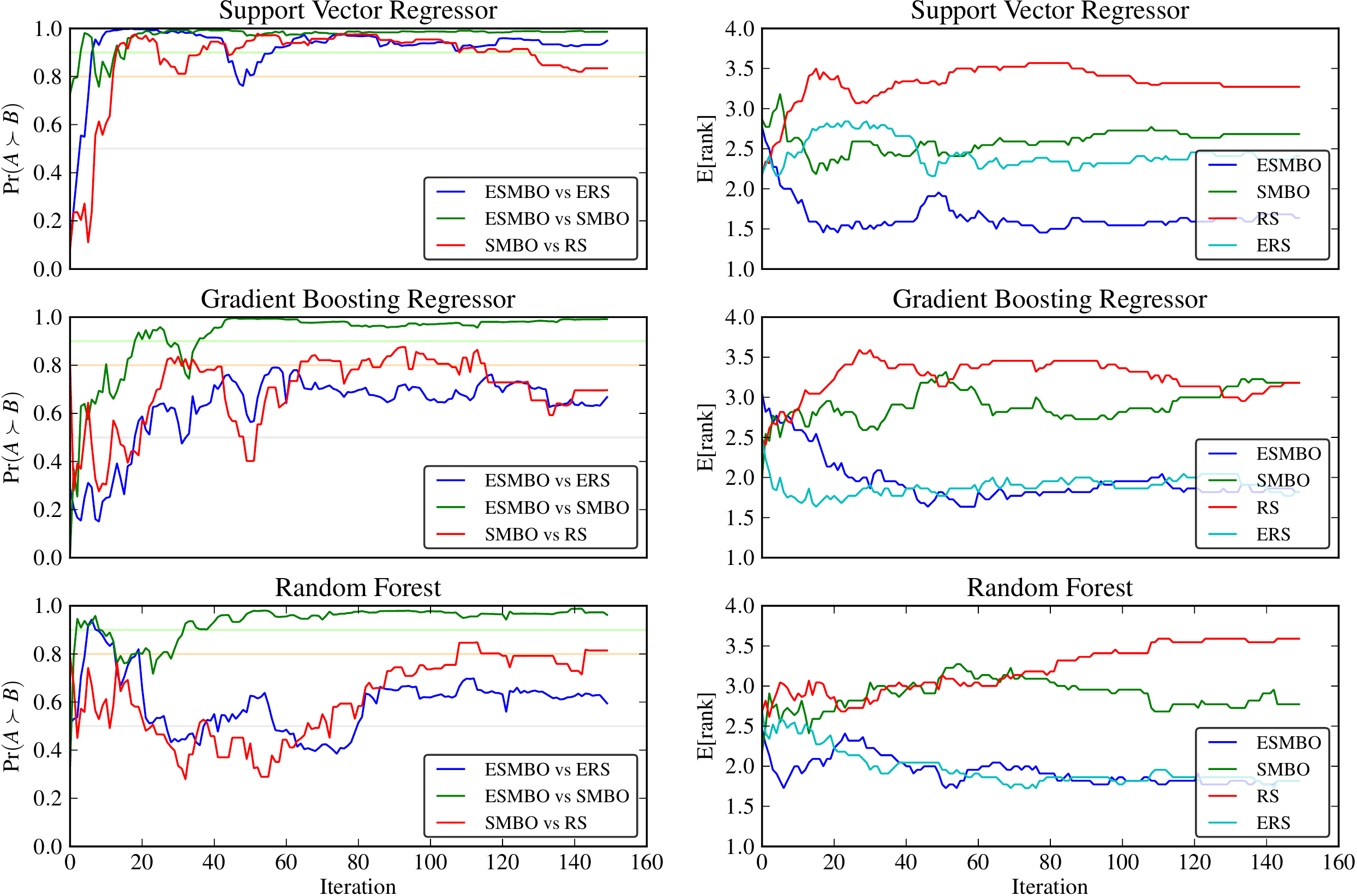}
\caption{PB probability and expected rankd over time for the 3 regression hyperparameter spaces.}
\label{fig:regression}
\end{center}
\end{figure*}

\begin{figure*}[t]
\begin{center}
\includegraphics[width=0.9\textwidth]{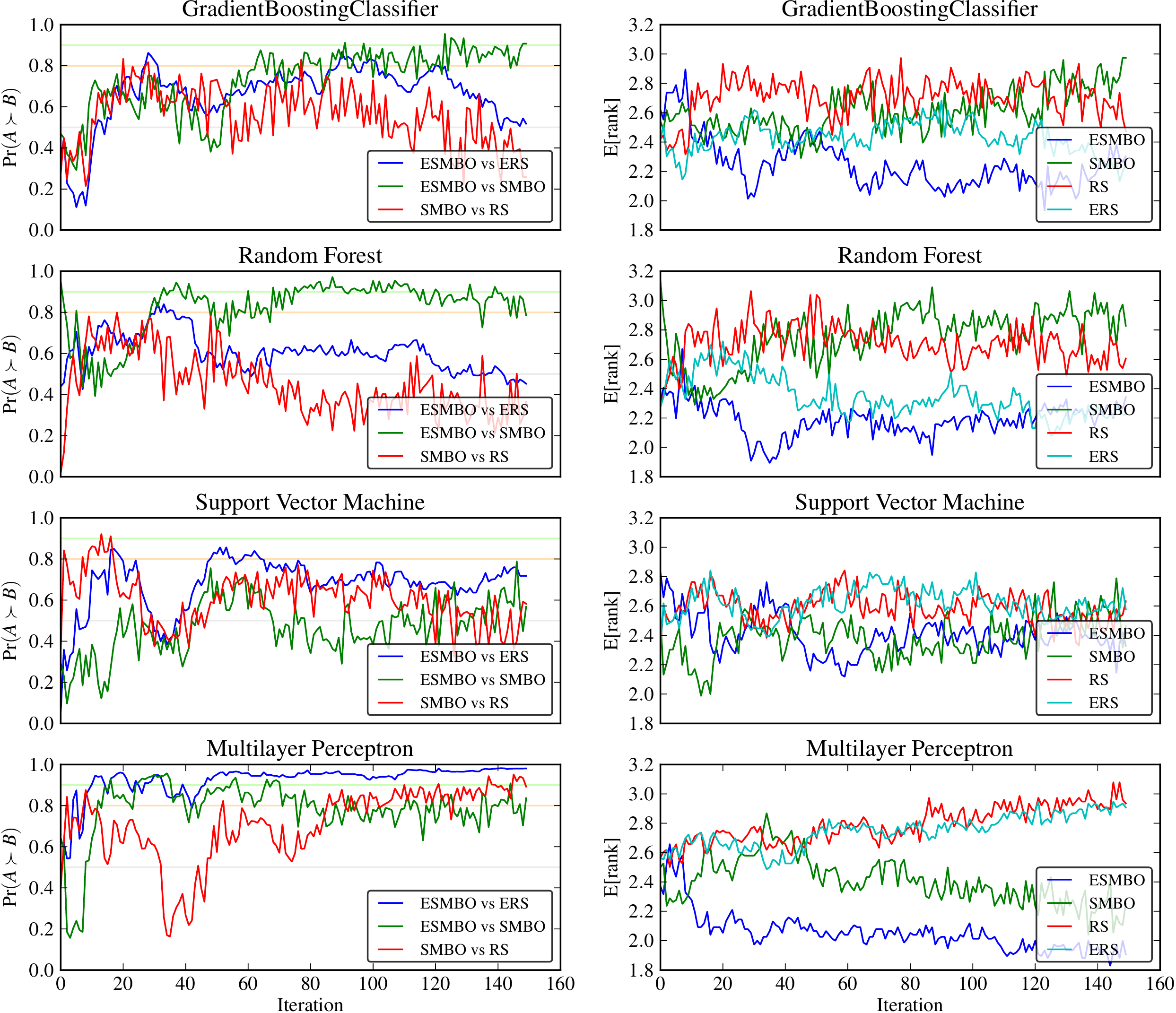}
\caption{PB probability and expected rankd over time for the 4 classification hyperparameter spaces.}
\label{fig:classification}
\end{center}
\end{figure*}

\providecommand{\stest}[2]{
    \hspace{-1.em}
    \textsuperscript{
            \textcolor[rgb]{#1}{\textbullet}\textcolor[rgb]{#2}{\textbullet}
    }
    \hspace{-1.em}
}

\providecommand{\NoSS}{0.95,0.95,0.95}
\providecommand{\LowSS}{1,0.5,0.0}
\providecommand{\HighSS}{0.1,0.8,0.1}
\providecommand{\gr}{\color{gray}}

\newcommand{\tabletitle}[1]{\vspace{3mm}\textbf{#1}}

\begin{table}[ht]
\caption{Pairwise win frequency for the 3 different regression hyperparameter spaces (Refer to Section~\ref{sec:table-notation} for the notation).}
\label{tab:regression}
\begin{center}

\begin{small}
\tabletitle{Support Vector Regressor}
\begin{tabular}{r|cccc||c}
      &             ESMBO             &             ERS             &              SMBO             &              RS             & E[rank] \\
\hline
ESMBO & \gr 0.50 \stest{\NoSS}{\NoSS} &  0.66 \stest{\HighSS}{\NoSS}  & 0.82 \stest{\HighSS}{\HighSS} & 0.86 \stest{\HighSS}{\HighSS} &   1.66  \\
ERS & \gr 0.34 \stest{\NoSS}{\NoSS} & \gr 0.50 \stest{\NoSS}{\NoSS} &   0.50 \stest{\NoSS}{\NoSS}   & 0.77 \stest{\HighSS}{\HighSS} &   2.38  \\
 SMBO & \gr 0.18 \stest{\NoSS}{\NoSS} & \gr 0.50 \stest{\NoSS}{\NoSS} & \gr 0.50 \stest{\NoSS}{\NoSS} &   0.64 \stest{\LowSS}{\NoSS}  &   2.68  \\
 RS & \gr 0.14 \stest{\NoSS}{\NoSS} & \gr 0.23 \stest{\NoSS}{\NoSS} & \gr 0.36 \stest{\NoSS}{\NoSS} & \gr 0.50 \stest{\NoSS}{\NoSS} &   3.27  \\
\end{tabular}

\tabletitle{Gradient Boosting Regressor}
\begin{tabular}{r|cccc||c}
      &             ERS             &             ESMBO             &              RS             &              SMBO             & E[rank] \\
\hline
ERS & \gr 0.50 \stest{\NoSS}{\NoSS} &   0.52 \stest{\NoSS}{\NoSS}   & 0.77 \stest{\HighSS}{\HighSS} & 0.86 \stest{\HighSS}{\HighSS} &   1.84  \\
ESMBO & \gr 0.48 \stest{\NoSS}{\NoSS} & \gr 0.50 \stest{\NoSS}{\NoSS} & 0.77 \stest{\HighSS}{\HighSS} & 0.91 \stest{\HighSS}{\HighSS} &   1.85  \\
 RS & \gr 0.23 \stest{\NoSS}{\NoSS} & \gr 0.23 \stest{\NoSS}{\NoSS} & \gr 0.50 \stest{\NoSS}{\NoSS} &   0.42 \stest{\NoSS}{\NoSS}   &   3.12  \\
 SMBO & \gr 0.14 \stest{\NoSS}{\NoSS} & \gr 0.09 \stest{\NoSS}{\NoSS} & \gr 0.58 \stest{\NoSS}{\NoSS} & \gr 0.50 \stest{\NoSS}{\NoSS} &   3.19  \\
\end{tabular}

\tabletitle{Random Forest}
\begin{tabular}{r|cccc||c}
      &             ESMBO             &             ERS             &              SMBO             &              RS             & E[rank] \\
\hline
ESMBO & \gr 0.50 \stest{\NoSS}{\NoSS} &   0.53 \stest{\NoSS}{\NoSS}   & 0.76 \stest{\HighSS}{\HighSS} & 0.91 \stest{\HighSS}{\HighSS} &   1.80  \\
ERS & \gr 0.47 \stest{\NoSS}{\NoSS} & \gr 0.50 \stest{\NoSS}{\NoSS} & 0.72 \stest{\HighSS}{\HighSS} & 1.00 \stest{\HighSS}{\HighSS} &   1.81  \\
 SMBO & \gr 0.24 \stest{\NoSS}{\NoSS} & \gr 0.28 \stest{\NoSS}{\NoSS} & \gr 0.50 \stest{\NoSS}{\NoSS} &   0.66 \stest{\NoSS}{\NoSS}   &   2.82  \\
 RS & \gr 0.09 \stest{\NoSS}{\NoSS} & \gr 0.00 \stest{\NoSS}{\NoSS} & \gr 0.34 \stest{\NoSS}{\NoSS} & \gr 0.50 \stest{\NoSS}{\NoSS} &   3.57  \\
\end{tabular}
\end{small}
\end{center}

\end{table}

\subsection{Table Notation}\label{sec:table-notation}

The result tables present the winning frequency for each pair of methods, where grayed out values represent redundant information. As a complement, we also add the expected rank of each method in the rightmost column and sort the table according to this metric. To report the conclusion of the Sign test and the PB test, we use colored dots, where orange means significant and green means highly significant. The first dot reports the result of the PB test and the second one, the Sign test. For more stable results, we average the values obtained during the last 15 iterations. 


\subsection{Analysis}

Looking at the overall results over 7 different hyperparameter spaces in Table~\ref{tab:regression} and Table~\ref{tab:classification}, we observe that ESMBO is never significantly outperformed by any other method and often outperforms the others. More precisely, it is either ranked first or tightly following ERS. Looking more closely, we see that the cases where ESMBO does not significantly outperform ERS concerns hyperparameter spaces of low complexity. For example, most hyperparameter configurations of Random Forest yield good generalization performances. Thus, these cases do not require an elaborate hyperparameter search method. On the other hand, when looking at more challenging hyperparameter spaces such as Support Vector Regression and Multilayer Perceptrons, we clearly see the benefits of combining SMBO with Agnostic Bayes. 

As described in Section~\ref{sec:our_method}, ESMBO is alternating between $N$ different SMBO optimizations and deviates from the natural sequence of SMBO. To see if this aspect of ESMBO can influence its convergence rate, we present a temporal analysis of the methods in Figure~\ref{fig:regression} and Figure~\ref{fig:classification}. The left columns depict $\Pr( A \succ B )$ for selected pairs of methods and the right columns present the expected rank of each method over time. 

A general analysis clearly shows that there is no significant degradation in terms of convergence speed. In fact, we generally observe the opposite. More precisely, looking at $\Pr( \operatorname{ESMBO} \succ \operatorname{SMBO} )$, the green curve of the left columns, it usually reaches a significantly better state right at the beginning or within the first few iterations. A notable exception to that trend occurs with the Multiplayer Perceptrons, where SMBO is significantly better than ESMBO for a few iterations at the beginning. Then, it gets quickly outperformed by ESMBO.

\begin{table}[ht]
\begin{center}
\begin{small}
\caption{Pairwise win frequency for the 3 different classification hyperparameter spaces (Refer to Section~\ref{sec:table-notation} for the notation). }
\label{tab:classification}

\tabletitle{Support Vector Machine}
\begin{tabular}{r|cccc||c}
      &             ESMBO             &              RS             &              SMBO             &             ERS             & E[rank] \\
\hline
ESMBO & \gr 0.50 \stest{\NoSS}{\NoSS} &   0.54 \stest{\NoSS}{\NoSS}   &   0.55 \stest{\NoSS}{\NoSS}   &   0.56 \stest{\NoSS}{\NoSS}   &   2.35  \\
 RS & \gr 0.46 \stest{\NoSS}{\NoSS} & \gr 0.50 \stest{\NoSS}{\NoSS} &   0.51 \stest{\NoSS}{\NoSS}   &   0.51 \stest{\NoSS}{\NoSS}   &   2.52  \\
 SMBO & \gr 0.45 \stest{\NoSS}{\NoSS} & \gr 0.49 \stest{\NoSS}{\NoSS} & \gr 0.50 \stest{\NoSS}{\NoSS} &   0.53 \stest{\NoSS}{\NoSS}   &   2.54  \\
ERS & \gr 0.44 \stest{\NoSS}{\NoSS} & \gr 0.49 \stest{\NoSS}{\NoSS} & \gr 0.47 \stest{\NoSS}{\NoSS} & \gr 0.50 \stest{\NoSS}{\NoSS} &   2.59  \\
\end{tabular}

\tabletitle{Gradient Boosting Classifier}
\begin{tabular}{r|cccc||c}
      &             ESMBO             &             ERS             &              RS             &              SMBO             & E[rank] \\
\hline
ESMBO & \gr 0.50 \stest{\NoSS}{\NoSS} &   0.51 \stest{\NoSS}{\NoSS}   &   0.59 \stest{\NoSS}{\NoSS}   &  0.65 \stest{\LowSS}{\HighSS} &   2.25  \\
ERS & \gr 0.49 \stest{\NoSS}{\NoSS} & \gr 0.50 \stest{\NoSS}{\NoSS} &   0.59 \stest{\NoSS}{\NoSS}   &  0.64 \stest{\LowSS}{\LowSS}  &   2.28  \\
 RS & \gr 0.41 \stest{\NoSS}{\NoSS} & \gr 0.41 \stest{\NoSS}{\NoSS} & \gr 0.50 \stest{\NoSS}{\NoSS} &   0.55 \stest{\NoSS}{\NoSS}   &   2.64  \\
 SMBO & \gr 0.35 \stest{\NoSS}{\NoSS} & \gr 0.36 \stest{\NoSS}{\NoSS} & \gr 0.45 \stest{\NoSS}{\NoSS} & \gr 0.50 \stest{\NoSS}{\NoSS} &   2.83  \\
\end{tabular}

\tabletitle{Random Forest}
\begin{tabular}{r|cccc||c}
      &             ERS             &             ESMBO             &              RS             &              SMBO             & E[rank] \\
\hline
ERS & \gr 0.50 \stest{\NoSS}{\NoSS} &   0.52 \stest{\NoSS}{\NoSS}   &   0.60 \stest{\NoSS}{\LowSS}  &  0.64 \stest{\LowSS}{\HighSS} &   2.24  \\
ESMBO & \gr 0.48 \stest{\NoSS}{\NoSS} & \gr 0.50 \stest{\NoSS}{\NoSS} &   0.60 \stest{\NoSS}{\NoSS}   &  0.67 \stest{\LowSS}{\HighSS} &   2.25  \\
 RS & \gr 0.40 \stest{\NoSS}{\NoSS} & \gr 0.40 \stest{\NoSS}{\NoSS} & \gr 0.50 \stest{\NoSS}{\NoSS} &   0.57 \stest{\NoSS}{\NoSS}   &   2.63  \\
 SMBO & \gr 0.36 \stest{\NoSS}{\NoSS} & \gr 0.33 \stest{\NoSS}{\NoSS} & \gr 0.43 \stest{\NoSS}{\NoSS} & \gr 0.50 \stest{\NoSS}{\NoSS} &   2.89  \\
\end{tabular}

\tabletitle{Multilayer Perceptron}
\begin{tabular}{r|cccc||c}
      &             ESMBO             &              SMBO             &             ERS             &              RS             & E[rank] \\
\hline
ESMBO & \gr 0.50 \stest{\NoSS}{\NoSS} &   0.57 \stest{\LowSS}{\NoSS}  & 0.76 \stest{\HighSS}{\HighSS} & 0.75 \stest{\HighSS}{\HighSS} &   1.92  \\
 SMBO & \gr 0.43 \stest{\NoSS}{\NoSS} & \gr 0.50 \stest{\NoSS}{\NoSS} &  0.68 \stest{\LowSS}{\HighSS} &  0.68 \stest{\LowSS}{\HighSS} &   2.21  \\
ERS & \gr 0.24 \stest{\NoSS}{\NoSS} & \gr 0.32 \stest{\NoSS}{\NoSS} & \gr 0.50 \stest{\NoSS}{\NoSS} &   0.54 \stest{\NoSS}{\NoSS}   &   2.91  \\
 RS & \gr 0.25 \stest{\NoSS}{\NoSS} & \gr 0.32 \stest{\NoSS}{\NoSS} & \gr 0.46 \stest{\NoSS}{\NoSS} & \gr 0.50 \stest{\NoSS}{\NoSS} &   2.96  \\
\end{tabular}
\end{small}
\end{center}

\end{table}


\section{Conclusion}

We described a successful method for automatically constructing ensembles without requiring hand-selection of models or a grid search. The method
can adapt the SMBO hyperparameter optimization algorithm so that it can produce an ensemble instead of a single model. Theoretically, the method is motivated by an Agnostic Bayesian paradigm which attempts to construct ensembles that reflect the uncertainty over which a model actually has the smallest true risk. The resulting method is easy to implement and comes with no extra computational cost at learning time. Its generalization performance and convergence speed are also dominant according to experiments on 22 regression and 39 classification data sets.

\clearpage

\bibliography{agnostic_Bayes}

\begin{thebibliography}{16}
\providecommand{\natexlab}[1]{#1}
\providecommand{\url}[1]{\texttt{#1}}
\expandafter\ifx\csname urlstyle\endcsname\relax
  \providecommand{\doi}[1]{doi: #1}\else
  \providecommand{\doi}{doi: \begingroup \urlstyle{rm}\Url}\fi

\bibitem[Bergstra et~al.(2011)Bergstra, Bardenet, Bengio, and
  K{\'e}gl]{bergstra2011}
James Bergstra, R{\'e}mi Bardenet, Yoshua Bengio, and Bal{\'a}zs K{\'e}gl.
\newblock Algorithms for hyper-parameter optimization.
\newblock In \emph{NIPS}, pages 2546--2554, 2011.

\bibitem[Hutter et~al.(2011)Hutter, Hoos, and
  Leyton-Brown]{hutter2011sequential}
Frank Hutter, Holger~H Hoos, and Kevin Leyton-Brown.
\newblock Sequential model-based optimization for general algorithm
  configuration.
\newblock In \emph{Learning and Intelligent Optimization}, pages 507--523.
  Springer, 2011.

\bibitem[Snoek et~al.(2012)Snoek, Larochelle, and Adams]{snoek2012practical}
Jasper Snoek, Hugo Larochelle, and Ryan~P. Adams.
\newblock Practical bayesian optimization of machine learning algorithms.
\newblock In \emph{NIPS}, pages 2960--2968, 2012.

\bibitem[Bell et~al.(2007)Bell, Koren, and Volinsky]{bell2007bellkor}
Robert~M Bell, Yehuda Koren, and Chris Volinsky.
\newblock The bellkor solution to the netflix prize.
\newblock \emph{KorBell Team's Report to Netflix}, 2007.

\bibitem[Lacoste et~al.(2014)Lacoste, Marchand, Laviolette, and
  Larochelle]{agnostic_bayes}
Alexandre Lacoste, Mario Marchand, Fran{\c{c}}ois Laviolette, and Hugo
  Larochelle.
\newblock Agnostic bayesian learning of ensembles.
\newblock In \emph{Proceedings of The 31st International Conference on Machine
  Learning}, pages 611--619, 2014.

\bibitem[Bergstra and Bengio(2012)]{bergstra2012randomhp}
James Bergstra and Yoshua Bengio.
\newblock Random search for hyper-parameter optimization.
\newblock \emph{Journal of Machine Learning Research}, 13:\penalty0 281--305,
  2012.

\bibitem[Laviolette et~al.(2011)Laviolette, Marchand, and
  Roy]{laviolette2011pac}
F.~Laviolette, M.~Marchand, and J.F. Roy.
\newblock From pac-bayes bounds to quadratic programs for majority votes.
\newblock \emph{moment}, 1500:\penalty0 Q2, 2011.

\bibitem[Kim and Ghahramani(2012)]{KimG12}
Hyun-Chul Kim and Zoubin Ghahramani.
\newblock Bayesian classifier combination.
\newblock \emph{Journal of Machine Learning Research - Proceedings Track},
  22:\penalty0 619--627, 2012.

\bibitem[Zhang et~al.(2006)Zhang, Burer, and Street]{zhang2006ensemble}
Yi~Zhang, Samuel Burer, and W~Nick Street.
\newblock Ensemble pruning via semi-definite programming.
\newblock \emph{The Journal of Machine Learning Research}, 7:\penalty0
  1315--1338, 2006.

\bibitem[Drucker et~al.(1997)Drucker, Burges, Kaufman, Smola, and
  Vapnik]{drucker1997support}
Harris Drucker, Chris~JC Burges, Linda Kaufman, Alex Smola, and Vladimir
  Vapnik.
\newblock Support vector regression machines.
\newblock \emph{Advances in neural information processing systems}, pages
  155--161, 1997.

\bibitem[Breiman(2001)]{breiman2001random}
L.~Breiman.
\newblock Random forests.
\newblock \emph{Machine learning}, 45\penalty0 (1):\penalty0 5--32, 2001.

\bibitem[Geurts et~al.(2006)Geurts, Ernst, and Wehenkel]{geurts2006extremely}
Pierre Geurts, Damien Ernst, and Louis Wehenkel.
\newblock Extremely randomized trees.
\newblock \emph{Machine learning}, 63\penalty0 (1):\penalty0 3--42, 2006.

\bibitem[Friedman(2001)]{friedman2001greedy}
Jerome~H Friedman.
\newblock Greedy function approximation: a gradient boosting machine.
\newblock \emph{Annals of Statistics}, pages 1189--1232, 2001.

\bibitem[Dem{\v{s}}ar(2006)]{demvsar2006statistical}
Janez Dem{\v{s}}ar.
\newblock Statistical comparisons of classifiers over multiple data sets.
\newblock \emph{The Journal of Machine Learning Research}, 7:\penalty0 1--30,
  2006.

\bibitem[Lacoste et~al.(2012)Lacoste, Laviolette, and
  Marchand]{lacostebayesian}
Alexandre Lacoste, Fran\c{c}ois Laviolette, and Mario Marchand.
\newblock Bayesian comparison of machine learning algorithms on single and
  multiple datasets.
\newblock \emph{Journal of Machine Learning Research - Proceedings Track},
  22:\penalty0 665--675, 2012.

\bibitem[Ula{\c s} et~al.(2009)Ula{\c s}, Semerci, Y{\i}ld{\i}z, and
  Alpayd{\i}n]{ulas2009}
Ayd{\i}n Ula{\c s}, Murat Semerci, Olcay~Taner Y{\i}ld{\i}z, and Ethem
  Alpayd{\i}n.
\newblock Incremental construction of classifier and discriminant ensembles.
\newblock \emph{Information Sciences}, 179\penalty0 (9):\penalty0 1298--1318,
  April 2009.

\end{thebibliography}
\bibliographystyle{unsrtnat}

\end{document}